\documentclass[10pt, a4paper]{article}
\usepackage{lrec}
\usepackage{multibib}
\newcites{languageresource}{Language Resources}
\usepackage{graphicx}
\usepackage{tabularx}
\usepackage{soul}
\usepackage{latexsym}
\usepackage{booktabs}
\usepackage{enumitem}
\usepackage[utf8]{inputenc}
\usepackage{hyperref}
\usepackage[usenames,dvipsnames,svgnames,table]{xcolor}
\usepackage{epstopdf}
\usepackage{xstring}

\hypersetup{
    colorlinks,
    citecolor=black,
    filecolor=black,
    linkcolor=black,
    urlcolor=black
}

\title{Baselines and Test Data for Cross-Lingual Inference}

\name{Željko Agić \quad Natalie Schluter}

\address{Department of Computer Science \\
IT University of Copenhagen \\
Rued Langgaards Vej 7, 2300 Copenhagen S, Denmark\\
\texttt{zeag@itu.dk \quad nael@itu.dk}}

\abstract{
The recent years have seen a revival of interest in textual entailment, sparked by i) the emergence of powerful deep neural network learners for natural language processing and ii) the timely development of large-scale evaluation datasets such as SNLI. Recast as natural language inference, the problem now amounts to detecting the relation between pairs of statements: they either contradict or entail one another, or they are mutually neutral. Current research in natural language inference is effectively exclusive to English. In this paper, we propose to advance the research in SNLI-style natural language inference toward multilingual evaluation. To that end, we provide test data for four major languages: Arabic, French, Spanish, and Russian. We experiment with a set of baselines. Our systems are based on cross-lingual word embeddings and machine translation. While our best system scores an average accuracy of just over 75\%, we focus largely on enabling further research in multilingual inference.
\\ \newline 
}

\begin{document}

\maketitleabstract

\section{Introduction}

Natural language processing is marking a very recent resurgence of interest in textual entailment. Now revamped as {\bf natural language inference} (NLI) by \newcite{bowman2015snli} with their SNLI dataset, the task of differentiating contradictory, entailing, and unrelated pairs of sentences (Fig.~\ref{fig:example}) has entertained a large number of proposals.\footnote{\scriptsize\url{https://nlp.stanford.edu/projects/snli/}} The timely challenge lends itself to various deep learning approaches such as by \newcite{rocktaschel2015reasoning}, \newcite{parikh2016decomposable}, or \newcite{wang2017bilateral}, which mark a string of very notable results.

Yet, the SNLI corpus is in English only. As of recently, it includes more test data from multiple genres,\footnote{\scriptsize\url{https://repeval2017.github.io/shared/}} but it remains exclusive to English. Following \newcite{bender2009naive} in seeking true language independence, we propose to extend the current NLI research beyond English, and further into the majority realm of low-resource languages.

Since training data is generally unavailable for most languages, work on transfer learning is abundant for the basic NLP tasks such as tagging and syntactic parsing \cite{das2011unsupervised,ammar2016many}. By contrast, the research in cross-lingual entailment is not as plentiful \cite{negri2013semeval}. To the best of our knowledge, at this point there are no contributions to SNLI-style cross-lingual inference, or for that matter, work on languages other than English at all.

\paragraph{Contributions.} In the absence of training data for languages other than English, we propose a set of baselines for {\bf cross-lingual neural inference}. We adapt to the target languages either by i) employing multilingual word embeddings or alternatively by ii) translating the input sentebces into English.

We create {\bf multilingual test data} to facilitate evaluation by manually translating  4 $\times$ 1,332 premise-hypothesis sentence pairs from the English SNLI test data into four other major languages: Arabic, French, Russian, and Spanish. We also experiment with automatic translations of the SNLI test data to serve as a proxy for large-scale evaluations in the absence of manually produced data.

\begin{figure}[t]
    \centering
	\resizebox{\columnwidth}{!}{
	\begin{tabular}{rl}
    \toprule
    \toprule
    \\[-2ex]
	{\bf premise} & Female gymnasts warm up before a competition. \\[1ex]
    {\bf entailment} & Gymnasts get ready for a competition.\\
	{\bf contradiction} & Football players practice.\\
	{\bf neutral} & Gymnasts get ready for the biggest competition of their life.\\[1ex]
    \bottomrule
	\end{tabular}
	}
	\caption{\label{fig:example}Example sentence 4-tuple from the SNLI test set, lines 758--760.}
\end{figure}

\section{Cross-Lingual Inference}

Following the success of neural networks in SNLI-style inference, we take the neural attention-based model of \newcite{parikh2016decomposable} as our starting point.  To date, their system remains competitive with the current state of the art.  As their attention model is based solely on word embeddings, and is independent of word order, it is particularly suitable for the baseline we present here: a purely multi-lingual embeddings based cross-lingual NLI system. Moreover, their approach is computationally much leaner than most competitors, making it a fast and scalable choice.\footnote{For more details, see the original paper, and an illustrative overview of the model: \scriptsize\url{https://explosion.ai/blog/deep-learning-formula-nlp.}}

In short, the \newcite{parikh2016decomposable} model sends sentence pairs, i.e., premises and hypotheses, through a neural pipeline that consists of three separate components:
\begin{itemize}[noitemsep,topsep=3pt,leftmargin=*]
\item[i)] \textbf{\textsc{attention:}} Scores combinations of pairs of words across input sentence pairs. Scores of these word pairs are given by a feed-forward network with ReLU activations that is assumed to model a homomorphic function for linear-time computation. Attention weights for phrases softly aligned with a word are obtained by summing their component vectors each factored by their normalized score.
\item[ii)] \textbf{\textsc{comparison:}} Word vectors and their aligned phrase counterparts are compared and combined into a single vector using a feed-forward neural network.
\item[iii)] \textbf{\textsc{concatenation:}} A network that sums over the above output vectors for each input sentence, concatenates this representation and feeds it through a final feed-forward network followed by a linear layer.
\end{itemize}
To be trained, the model expects SNLI annotations, and an ideally very large vocabulary of distributed word representations.

In this paper, we have at our disposal only a large training corpus of English NLI examples, but a distinct language in which we want to predict for NLI: the target language. We train the system described above on the English training set. We exploit the fact that the system is purely embeddings-based and train with multilingual embeddings for a set of languages including English and the prediction language. Multilingual embeddings are sets of word embeddings generated for multiple languages where the embeddings from the union of these sets are meant to correspond to one another semantically independent of the language the words the embeddings correspond to actually belong. At prediction time, we can safely use the embeddings of the target language.

\paragraph{Mapping.} One method for obtaining multilingual word embeddings is to apply the translation matrix technique to a set of monolingual embeddings \cite{mikolov2013exploiting} with the aid of a bilingual dictionary containing the source-target word pairs. The method works by finding a transformation matrix from the target language monolingual embeddings to the English monolingual embeddings that minimizes the total least-squared error. This transformation matrix can then be used on words not seen in the bilingual dictionary.

\paragraph{Multilingual embeddings.} If parallel sentences or even just parallel documents are available for two or more languages, we can use this data to embed their vocabularies in a shared representation. For example, through an English-Russian parallel corpus we would represent the words of the two languages in a shared space.

There are several competing approaches to training word embeddings over parallel sentences. In this paper, we experiment with four.

\begin{itemize}[noitemsep,topsep=3pt,leftmargin=0pt]
\item[] \textbf{\textsc{bicvm:}} The seminal approach by \newcite{hermann2014multilingual} for inducing bilingual compositional representations from sentence-aligned parallel corpora only.\footnote{\scriptsize\url{https://github.com/karlmoritz/bicvm}}
\item[] \textbf{\textsc{invert:}} Inverted indexing over parallel corpus sentence IDs as indexing features, with SVD dimensionality reduction on top, following \newcite{sogaard2015inverted} in the recent implementation by \newcite{levy2017strong}.\footnote{\scriptsize\url{https://bitbucket.org/omerlevy/xling_embeddings/}} Instead of embedding just language pairs, this method embeds multiple languages into the same space. It is thus distinctly multilingual, rather than just bilingual.
\item[] \textbf{\textsc{random:}} Our implementation of the approach by \newcite{vulic2016bilingual} whereby bilingual SGNS embeddings of  \newcite{mikolov2013distributed} are trained on top of merged pairs of parallel sentences with randomly shuffled tokens.
\item[] \textbf{\textsc{ratio:}} Similar to {\sc random}, except the tokens in bilingual sentences are not shuffled, but inserted successively by following the token ratio between the two sentences.
\end{itemize}

\paragraph{Machine translation.}

One alternative to adapting via shared distributed representations is to use machine translation.

If high-quality translation systems are readily available, or if we can build them from abundant parallel corpora, we can simply translate any input to English and run a pre-trained English NLI model over it. Moreover, we can translate the training data and train target language models similar to \newcite{tiedemann2014treebank} in cross-lingual dependency parsing.

The MT approach only lends itself to medium- to high-density languages. The mapping requires only the monolingual data and bilingual dictionaries, while the bilingual embeddings need parallel texts or documents, both of which are feasible for true low-resource languages.

\begin{table}
\centering
\begin{tabular}{rcccc}
\toprule
& {\bf ara} & {\bf fra} & {\bf spa} & {\bf rus} \\
{\bf eng} to ... & 25.58 & 55.80 & 39.65 & 30.31 \\
... to {\bf eng} & 37.48 & 46.90 & 44.04 & 31.17 \\
\bottomrule
\end{tabular}
\caption{\label{tbl:bleu} Machine translation quality (BLEU) for translating the test data from and into English.}
\end{table}

\section{Test Data}


The SNLI data are essentially pairs of sentences---premises and hypotheses---each paired with a relation label: contradiction, entailment, or neutral. We had human experts manually translate the first 1,332 test pairs from English into Arabic, French, Russian, and Spanish. We copied over the original labeling of relations, and the annotators manually verified that they hold. That way we can directly evaluate the NLI performance for these five languages.

Further, we translated our test sets into English by Google Translate for our MT-based system as it adapts through translation and thus expects input in English. We also automatically translated the 1,332 original English sentences into our new test languages to check how well we can approximate the ``true'' accuracies by using translated test data. This way we can facilitate cross-lingual NLI evaluations on a larger scale.

The BLEU scores for the two translation directions are given in Table~\ref{tbl:bleu}, where we see a clear split by similarity as the translations tend to be better between English, French, and Spanish, and worse outside that group.

\section{Experiment}

Our experiment involves adapting a neural NLI classifier through multilingual word embeddings and machine translation. We run the \newcite{kim2017structured} implementation of the attention-based system of \newcite{parikh2016decomposable}.\footnote{\scriptsize\url{https://github.com/harvardnlp/struct-attn}} All models are trained for 15 epochs and otherwise with default settings. While this system typically peaks at over 100 epochs, we sacrifice some accuracy to provide more data points in the comparison given the time constraints.

We set the dimensionality to 300 for all our embeddings. Other than that, they are trained with their default settings. In mapping we use the pretrained {\sc fasttext} vectors\footnote{\scriptsize\url{https://github.com/facebookresearch/fastText/blob/master/pretrained-vectors.md}} for all five languages \cite{bojanowski2016enriching}. We map the target language embeddings to English as \newcite{mikolov2013exploiting}, using the \newcite{dinu2014improving} implementation\footnote{\scriptsize\url{http://clic.cimec.unitn.it/~georgiana.dinu/down/}} and Wiktionary data.\footnote{\scriptsize\url{https://dumps.wikimedia.org/}}

We train our bilingual embeddings on the UN corpus \cite{ziemski2016united}. The corpus covers English and the four target languages with 11M sentences each. The sentences are aligned across all five languages. The Moses tokenizer\footnote{\scriptsize\url{https://github.com/moses-smt/mosesdecoder/}} \cite{koehn2007moses} was used to preprocess the corpus and the test data for training and evaluation.

In the MT approach, we only experiment with translating the input, and not with translating the training data due to time constraints. There, we use two English SNLI models: one with {\sc fasttext} and the other with {\sc glove} 840B embeddings \cite{pennington2014glove}.\footnote{\scriptsize\url{https://nlp.stanford.edu/projects/glove/}}

\paragraph{Results.}

\begin{table}
\centering
\resizebox{\columnwidth}{!}{
\begin{tabular}{rccccc}
\toprule
& {\bf ara} & {\bf eng} & {\bf fra} & {\bf spa} & {\bf rus} \\
{\bf map to eng} \\
{\sc fasttext} &  55.75 & \textcolor{lightgray}{79.74} & 51.64 & 51.94 & 48.59 \\[1ex]
{\bf bilingual} \\
{\sc bicvm} &  56.82 & \textcolor{lightgray}{76.26} & 59.03 & 59.48 & 54.30 \\
{\sc random} &  57.35 & \textcolor{lightgray}{77.42} & 63.21 & 61.01 & 56.97 \\
{\sc ratio} &  54.46 & \textcolor{lightgray}{78.10} & 58.64 & 60.09 & 51.18 \\[1ex]
{\bf multilingual} \\
{\sc invert}  &  54.76 & \textcolor{lightgray}{75.10} & 62.60 & 60.55 & 54.76 \\
[1ex]
{\bf translation} \\
{\sc fasttext} &  72.28 & \textcolor{lightgray}{--}  & 77.23 & 75.93 & 76.54 \\
{\sc glove} & 75.86 & \textcolor{lightgray}{--} & 80.05 & 78.75 & 79.59\\
\bottomrule
\end{tabular}
}
\caption{\label{tbl:results} Overall accuracy of the cross-lingual approaches for the target languages \textcolor{lightgray}{and English}.}
\end{table}

We report the overall accuracy and F$_1$ scores for the three labels. Table~\ref{tbl:results} gives the overall scores of our cross-lingual NLI approaches. In general, the more resources we have, the better the scores: Training bilingual embeddings surpasses the mapping to English, while translating to English using a top-level MT system tops the adaptation via embeddings.

The mapping to English works slightly better for Arabic than for the other languages, and scores an average of 52\%. The {\sc random} bilingual embeddings top their group with an average accuracy of 59.6\% followed by {\sc invert} at 58.1\%, while {\sc ratio} and {\sc bicvm} are below at 56.1 and 57.4\%. The MT approach expectedly tops the table at 75.5\% accuracy. In Table~\ref{tbl:classes} we see that our best bilingual embeddings system {\sc random} has a preference for entailment, with ca 9\% in F$_1$ over the other two labels, which makes sense for a model aimed at capturing semantic similarity. This also holds true for the original \newcite{parikh2016decomposable} evaluation on English.


We report all our English scores as a sanity check. In 100 training epochs, \newcite{parikh2016decomposable} score 86.8\% with {\sc glove} 840B as their top score, while we mark 83.4\% in 15 epochs. With the significantly smaller {\sc fasttext} embeddings we reach an accuracy of 79.7\%. The multilingual embeddings average at 76.7\% for English, where {\sc ratio} peaks at 78.1\%, likely as its sequential shuffling of parallel texts most closely captures the English sentence structure.

\paragraph{Discussion.}

\begin{table}[t]
\centering
\begin{tabular}{rcccc}
\toprule
& {\bf con} & {\bf ent} & {\bf neu} \\
{\bf ara} & 55.82 & 64.17 & 50.91 \\
{\bf fra} & 57.63 & 68.73 & 61.72 \\
{\bf spa} & 55.78 & 66.98 & 57.80 \\
{\bf rus} & 56.83 & 60.61 & 53.29 \\
\bottomrule
\end{tabular}
\caption{\label{tbl:classes} F$_1$ scores for {\bf con}tradiction, {\bf ent}ailment, and {\bf neu}tral for our best system, {\sc random}.}
\end{table}

\begin{figure}[t]
    \centering
    \includegraphics[width=0.85\columnwidth]{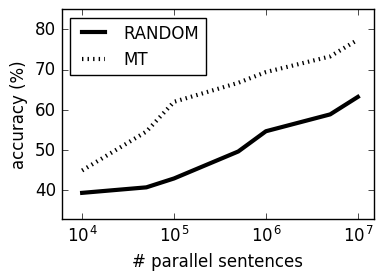}
    \caption{\label{fig:curve}French NLI accuracy in relation to parallel corpus size for {\sc random} embeddings.}
\end{figure}

Figure~\ref{fig:curve} plots a learning curve for the French {\sc random} approach. We see that its accuracy steadily increases by adding more parallel data into building the bilingual embeddings. As a side note, the MT-based system benefits if the English side of the embeddings grows in size and quality. The figure points out that i) adding more data benefits the task, and that ii) the accuracy of our {\sc random} approach stabilizes at around 1M parallel sentences. As per \newcite{sogaard2015inverted} most language pairs can offer no more than 100k sentence pairs, this puts forth a challenge for future cross-lingual NLI learning research.

Replacing the manually prepared test sets with the ones automatically translated from English underestimates the true accuracy by absolute -2.57\% on average. The higher the translation quality, the better the estimates we observe: While the difference is around -1\% for French and Spanish, it is -7\% for Arabic. Still, in proxy evaluation, as with our MT-based adaptation approach in general, we exercise caution: SNLI sentences are image captions, mostly $\leq$15 words long and thus relatively easy to translate (cf. \newcite{bowman2015snli}, Fig.~2) in comparison to, e.g., newspaper text.

\section{Related Work}

Prior to SNLI, there has been work in cross-lingual textual entailment using parallel corpora \cite{mehdad2011using} and lexical resources \cite{castillo2011wordnet}, or crowdsourcing for multilingual training data by \newcite{negri2011divide}. We also note two shared tasks, on cross-lingual entailment with five languages \cite{negri2013semeval} and English relatedness and inference \cite{marelli2014semeval}.

\newcite{cer2017semeval} provide multilingual evaluation data within a shared task in semantic textual similarity. There, paired snippets of text are evaluated for their degree of equivalence, and could thus be treated as a fine-grained proxy for SNLI-style evaluations.

SNLI is the first large-scale dataset for NLI in English \cite{bowman2015snli}, two orders of magnitude larger than any predecessor. It was recently expanded with test data for multiple genres of English to allow for cross-domain evaluation.\footnote{\scriptsize\url{https://www.nyu.edu/projects/bowman/multinli/}} Prior to our work, there have been no SNLI-style cross-lingual methods or evaluations.

\section{Conclusions}

We have proposed the first set of cross-lingual approaches to natural language inference, together with novel test data for four major languages. In experiments with three types of transfer systems, we record viable scores, while at the same time exploring the scalability of cross-lingual inference for low-resource languages.

We are actively enlarging the test data and introducing new languages. Our multilingual test sets and word embeddings are freely available.\footnote{\scriptsize\url{https://bitbucket.org/nlpitu/xnli}}



\section{References}

\bibliographystyle{lrec}
\bibliography{xample}


\end{document}